\documentclass[sn-mathphys]{sn-jnl}% Math and Physical Sciences Reference Style
\usepackage{xcolor}
\usepackage{appendix}
\usepackage{amsmath}
\usepackage{graphicx}
\usepackage{lineno}
\usepackage{array}
\usepackage{longtable}
\usepackage{natbib}
\usepackage{setspace}
\usepackage{color, colortbl}
\usepackage{amssymb}
\usepackage{amsmath}
\usepackage{graphicx}
\usepackage{amsfonts}       % blackboard math symbols
\usepackage{nicefrac}       % compact symbols for 1/2, etc.
\usepackage{microtype}      % microtypography
\usepackage{lipsum}		% Can be removed after putting your text content
\usepackage{graphicx}
\usepackage{natbib}
\usepackage{doi}
\usepackage{setspace}
\usepackage{ragged2e}
\usepackage{geometry}

\newcommand\TstrutNorm{\rule{0pt}{2.6ex}}         % = `top' strut
\newcommand\BstrutNorm{\rule[-1ex]{0pt}{0pt}}   % = `bottom' strut

\theoremstyle{thmstyleone}%
\theoremstyle{thmstyletwo}%
\theoremstyle{thmstylethree}%

\begin{document}
	
\title[Recent Trends in 2D Obj. Detection and Appl. in Video Event Recognition]{Recent Trends in 2D Object Detection and Applications in Video Event Recognition}

\author[1]{Prithwish Jana}\email{pjana@ieee.org}
\author[2]{Partha Pratim Mohanta}\email{partha.p.mohanta@gmail.com}

\affil[1]{Department of Computer Science \& Engineering, \\ Indian Institute of Technology Kharagpur}
\affil[2]{Electronics \& Communication Sciences Unit, \\ Indian Statistical Institute, Kolkata}

\abstract{Object detection serves as a significant step in improving performance of complex downstream computer vision tasks. It has been extensively studied for many years now and current state-of-the-art 2D object detection techniques proffer superlative results even in complex images. In this chapter, we discuss the geometry-based pioneering works in object detection, followed by the recent breakthroughs that employ deep learning. Some of these use a monolithic architecture that takes a RGB image as input and passes it to a feed-forward ConvNet or vision Transformer. These methods, thereby predict class-probability and bounding-box coordinates, all in a single unified pipeline. Two-stage architectures on the other hand, first generate region proposals and then feed it to a CNN to extract features and predict object category and bounding-box. We also elaborate upon the applications of object detection in video event recognition, to achieve better fine-grained video classification performance. Further, we highlight recent datasets for 2D object detection both in images and videos, and present a comparative performance summary of various state-of-the-art object detection techniques.}


\maketitle

\section{Introduction}
\begin{figure}[h!]
	\includegraphics[width=\linewidth]{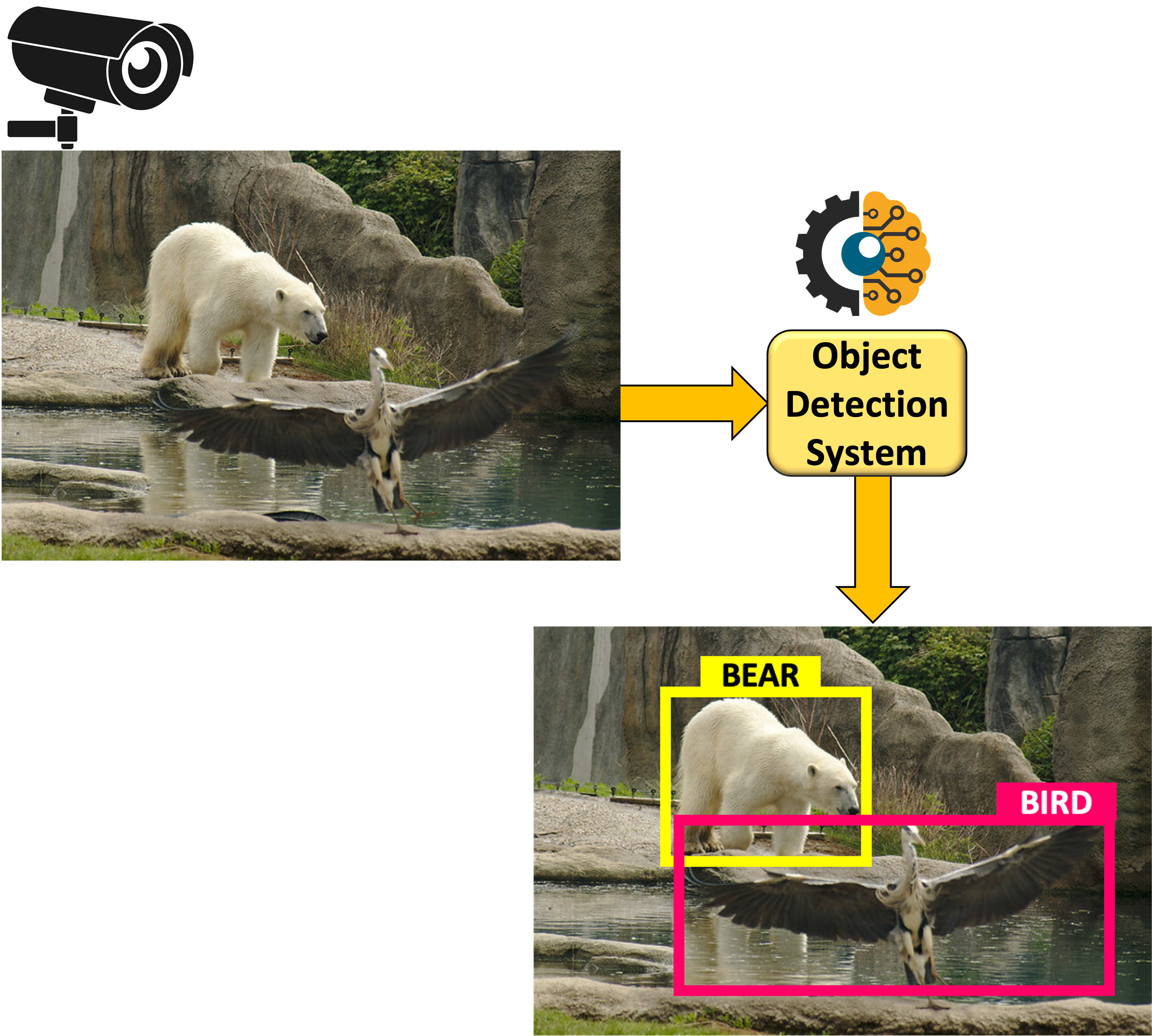}
	\caption{Digital images are captured by cameras through perception subsystem. The example image (from MS COCO dataset~\cite{lin2014microsoft}) when fed to an object detection system, detects two objects, estimates tight bounding boxes around them and classifies the object within as `bear' and `bird'.}
	\label{fig:objectDetectionIntro_bearBird_vert}
\end{figure}

Computer vision deals with mimicking the human visual perception system, such that computers can `see' (perceive and understand) their surrounding visual scenes the same way as we humans do. Through their perception subsystem, computers first perceive scenes through cameras, which are sent for internal processing commonly in the form of images and videos. One of the important tasks on images is \textit{object detection}, and it may be regarded as a primary step in many computer vision pipelines to support downstream complex tasks like panoptic and instance segmentation, image captioning, etc. Object detection deals with identifying instances of certain entities (e.g. human, dog, chair, tree, etc.) within a digital image. As illustrated in Figure~\ref{fig:objectDetectionIntro_bearBird_vert}, Identification is in terms of \textit{localizing} a bounding box which encompasses an object and \textit{classifying} the object present within the bounding box. 

\begin{figure}[h!]
	\includegraphics[width=\linewidth]{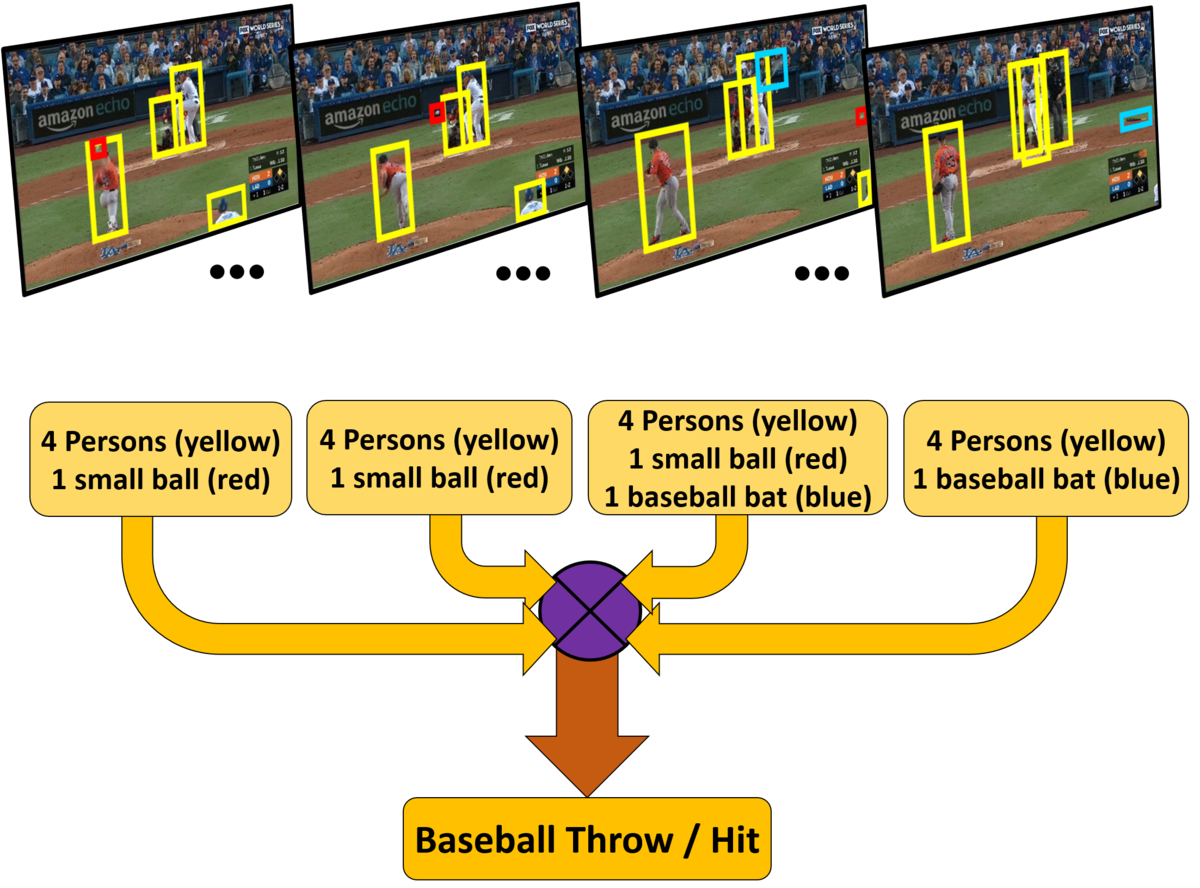}
	\caption{Object detection on each frame of a video gives an estimate of the relative position of each entity throughout the video duration. Efficient combination of this sequential information can lead to superlative event recognition in videos. For the example video~\cite{piergiovanni2018fine}, detected objects (person, baseball bat, ball) are shown per-frame, that are used to predict final event `Baseball Throw / Hit'}
	\label{fig:objDetInVids_baseball}
\end{figure}

The task of video event recognition~\cite{bhaumik2021event} is to predict the ongoing event or activity in a video throughout its duration. The prevalent approach is to obtain a global video-level feature representation~\cite{jana2021unsupervised} from 3D CNNs and classify the video using the same. However, this is generally insufficient for fine-grained tasks. As for example, it is a difficult feat~\cite{jana2019key} to discriminate similar events like anniversary event and birthday party, both of which showcase some kind of gathering of people and celebration. In such cases, frame-wise object detection may serve to be useful that can identify objects involved throughout the video duration. The overall idea is depicted in Figure~\ref{fig:objDetInVids_baseball}. Event-specific objects (e.g. balloons, toys) or event-specific inter-frame motion of objects (e.g. periodic round-about motion of persons can hint about the game of musical chairs) can be particularly useful in fine-grained discrimination of video events. 

The rest of the chapter is organized as follows. In Section~\ref{relwork} of this chapter, we first highlight some of the early geometric approaches to object detection. We move on to describe some of the recent major advancements in object detection that employ deep learning approaches. Next, in Section~\ref{vidActbyObjDet}, we elucidate on techniques of fine-grained video event recognition by frame-wise object detection. In Section~\ref{datasets}, we present some of the 2D object recognition datasets both in the domain of images and videos. We compare some of the recent state-of-the-art object detection techniques in Section~\ref{perfComparison}. Finally, we conclude in Section~\ref{epilogue}.

\section{Related Work on 2D Object Detection}
\label{relwork}

In this section, we discuss some of the early rule-based and geometry-based 2D object detection techniques. We further describe later progresses made by adopting deep learning based architectures that employ either monolithic architectures (a single stage object detection pipeline) and two-stage architectures (where region proposals are generated first, followed by classification, regression and other post-processing). 

\subsection{Early Object Detection: Geometrical \& Shape-based Approaches}
The earliest object detection techniques were majorly based on pattern or template matching. One such notable method was that by Fischler et al.'s~\cite{fischler1973representation} in 1973, who devised a technique of matching pictorial structures. The authors mention a commonality between different images of the same object, viz. they all can be broken down into a number of fundamental \textit{structures} possessing some preset relative spatial position. As such, given a test image, if all such fundamental structures are present and they maintain the same relative position it can be said that the object is present within the image. They put forward an embedding metric to determine how well each of the fundamental structures make up the total composite object that is being probed and used sequential optimization for matching. Most of the object detection methods in that era exploited object shape and geometry to solve the detection problem. Such approaches were in general `hard-detection' techniques, where a composite shape and inter-structure configuration is evaluated only after all constituent structures could be recognized and detected in an image. In contrast to such hard-detection, Burl et al.~\cite{burl1998probabilistic} suggested a soft-detection technique to identify an object composed of several structures whose relative position may vary upto an extent. As a part of this, they probed that particular relative configuration of these structures which simultaneously maximized the composite shape's log-likelihood (global criteria) and matching score for the structures (local criteria). The scenario becomes somewhat different when images are not in RGB format, but is grayscale (i.e. one channel, e.g. infrared images used in military operations) or stacked (e.g. confocal microscopy images). Quy et al.~\cite{quy2014using} and Miao et al.~\cite{miao2021real} use binarization~\cite{jana2017fuzzy} to perform thresholding on the grayscale infrared images and compute bounding-box around the prominent-cum-salient connected components. Luo et al.~\cite{luo2018sensitive} also use thresholding as a preprocessing step before fine-grained object detection in confocal laser-scanning microscope image stacks. As time progressed, the trend of shape-based template matching and rule-based methods for object detection shifted towards statistical approaches and machine learning-based classifiers.  

\subsection{Modern Object Detection Techniques: Use of Deep Learning}

We discuss some of the recent significant monolithic and two-stage architectures for 2D object detection, in the subsequent paragraphs. 

\subsubsection{Monolithic Architectures for 2D Object Detection}

The work on \textit{DetectorNet} by Szegedy et al.~\cite{szegedy2013deep} was one of the earliest that incorporated deep neural networks to predict category and precise location of objects within an image. Prior to that, deep convolutional neural networks (CNN) were mostly used for image classification tasks. On the contrary, DetectorNet was built upon the AlexNet~\cite{krizhevsky2012imagenet} architecture which was specialized for the task of detection. This was done by replacing the last layer of softmax with a regression layer that outputs a square-shaped mask of fixed size. They considered input of receptive field $225\times 225$ and output mask of size $24\times 24$. Further, they dissect the image into multi-scale coarse grids and consider four supplementary networks for estimating the top, bottom, right and left halves of objects in these grids. All these predicted masks on coarse grid help each other and the full mask in correcting mistakes and on combining and refinement, proffers the predicted bounding box for objects. To train the network, $L2$-loss of the predicted mask and the ground-truth mask is minimized over all samples in the training set. Although this course-to-fine approach of object mask regression was a pioneer in CNN-based object detection, but there was a disadvantage. Since the whole CNN architecture needs to be trained separately for every object category and every mask type, this incurs huge computational complexity.

Another notable work published at around the same time was \textit{OverFeat}, put forward by Sermanet et al.~\cite{Sermanet2014OverFeatIR}. Similar to DetectorNet, this is also a monolithic shared architecture to simultaneously learn for classification, localization and detection of objects. By an efficient sliding-window approach, the network is run across each image location and for multiple scales such that the computations for overlapping regions are done only once. This way, OverFeat is much speedier than DetectorNet. Next, the classification layer of softmax is once again replaced by a regression layer and thereafter it is trained to predict bounding boxes. The classifier and the regressor networks are sequentially trained to generate probable class label, its confidence (from softmax output) and the horizontal/vertical shifts of each bounding-box proposals.

Single-shot Multibox Detector (SSD) by Liu et al.~\cite{liu2016ssd} used VGG-16 architecture as a backbone because it performed the best in image classification and transfer learning tasks at that time. It is a single-shot object detection model. In place of the fully-connected layers at the bottom of VGG-16, the authors introduced multiple auxiliary convolutional feature layers. Each auxiliary layer is responsible for decreasing the hidden layer output progressively to different output sizes. This is used to estimate the confidence scores and offset values to varied scales of default boxes possessing disparate aspect ratios. As such, using multi-scale feature maps and convolutional kernels for the task of detection, SSD could proffer high accuracy with highly-improved computation speed. Feature Fusion SSD (FSSD)~\cite{li2017fssd} was an improvement on top of SSD to include a lightweight add-on module for the purpose of fusing features. In SSD, features pertaining to various scales could not be fused efficiently because it used feature pyramid detection. This was improved in FSSD where, features from different layers pertaining to separate scales were concatenated together by feature fusion, to generate new feature pyramids thereafter. This gives a rise in accuracy but a drop in the detection speed, when compared to SSD. Attentive SSD (ASSD)~\cite{yi2019assd} uses a scheme of incorporating attention whereby, significant regions on feature maps were preserved and emphasized while doing away with unnecessary regions. 

\begin{figure}[h!]
	\includegraphics[width=0.97\linewidth]{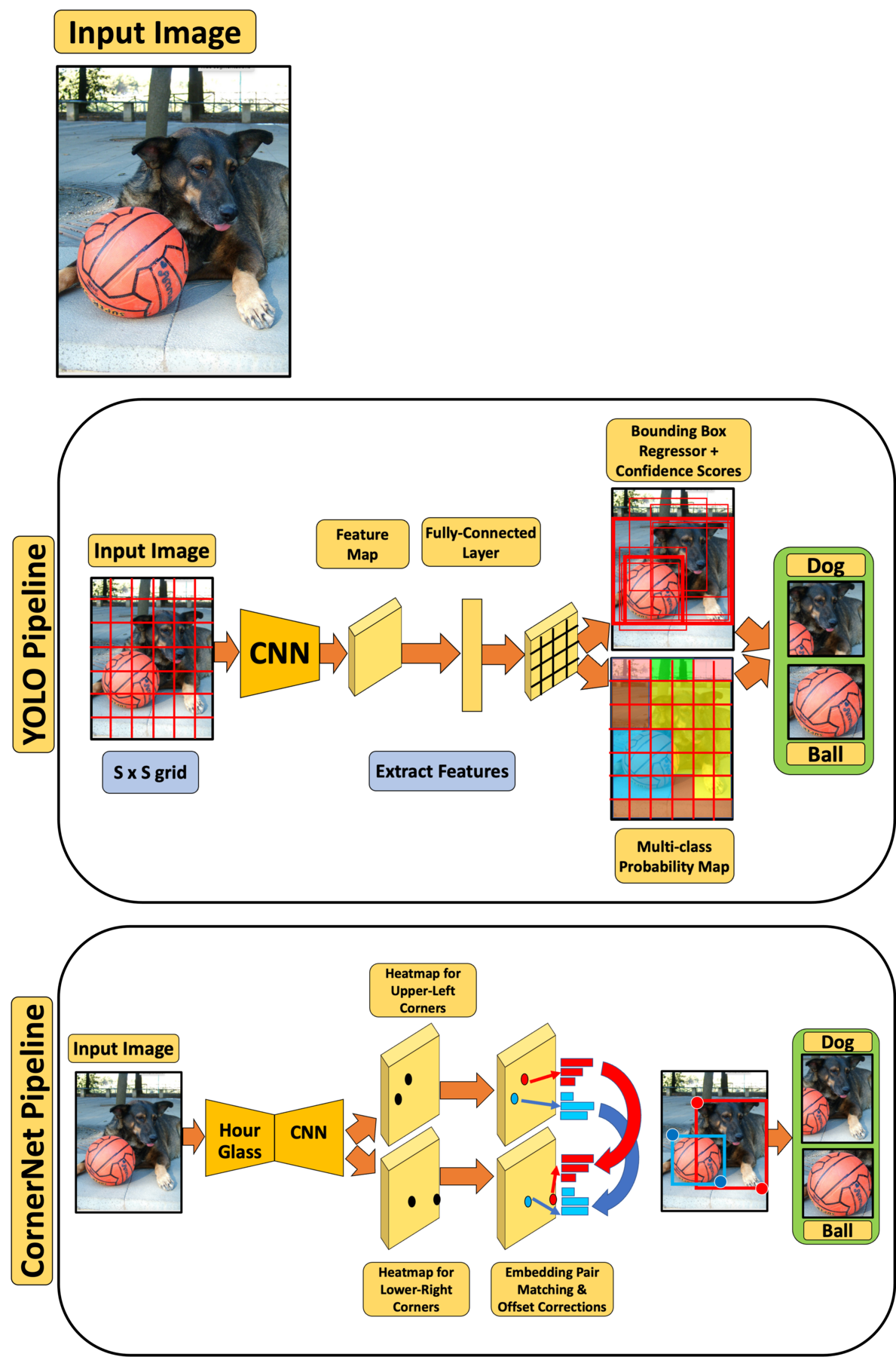}
	\caption{Comparison of YOLO and CornerNet pipeline (both monolithic architectures) for an image from COCO~\cite{lin2014microsoft}}
	\label{fig:yolo_cornernet_pipeline}
\end{figure}

Redmon et al.~\cite{redmon2016you} came up with another significant breakthrough in the domain of single-shot 2D object detection. Unlike many earlier approaches, this model entitled \textit{You Only Look Once (YOLO)}, was not a classifier disguised and modified for object detection task. Moreover, this was capable of detecting objects in real-time speed. Its overall framework is shown in as illustrated in Figure~\ref{fig:yolo_cornernet_pipeline}. Here, the input image is first partitioned into a grid each of whose cell is $S\times S$. Each such cell has the responsibility of classifying and localizing an object whose center lies within that particular cell. A bounding box prediction is defined by five values, viz. $x, y, w, h, CF$ where $x,y$ represents the spatial coordinate of the box center, $w,h$ determines the height and width of that box and $CF$ (objectness score) is a measure of how confident the architecture is about presence of an object within that cell. The confidence score is obtained by Intersection-over-Union ratio with a ground-truth bounding box and for overlapping proposals, only the proposal with highest $CF$ is preserved to be computationally efficient. Finally, each box proposal is separately classified to get probability over all classes. This class-wise probability and the objectness score together gives a score of how possible it is for an object of particular class to be present within the bounding-box proposal. But, YOLO has a disadvantage that it cannot predict small objects located at close proximity because each cell can only have two bounding box predictions of a single class. Many improvements over this came in the subsequent years. One such approach was YOLOv2~\cite{redmon2017yolo9000} that catered to improving recall and localization in YOLO by changes like introducing batch normalization, increasing input resolution, prediction of multiple objects per grid-cell with anchors, etc. Fast YOLO~\cite{shafiee2017fast} improved upon YOLOv2 architecture where motion-adaptive inference was employed to attain even faster detection with a more compact architecture. YOLOv3~\cite{farhadi2018yolov3} computes objectness score for each bounding box by logistic regression and uses a relatively larger architecture DarkNet-19 with residual shortcut connections. As such, it shows much improved performance on small close-by objects as compared to YOLO. YOLOv4~\cite{bochkovskiy2020yolov4} is a more faster alternative by incorporating bag-of-freebies and bag-of-freebies techniques during the training phase.

Other popular one-stage object detection frameworks include CornerNet~\cite{law2018cornernet} which does not necessitate formulation of good anchor boxes. Using a single CNN pipeline, they predict separate heatmaps for upper-left and lower-right corners of all object instances that belong to the same class. Further, the CNN is trained to predict an embedding vector with each such corner points, such that corresponding pair of upper-left and lower-right corners for the same objects possess a minimal inter-embedding distance. Such pairs are thereby matched (as illustrated in Figure~\ref{fig:yolo_cornernet_pipeline}) to form a bounding-box encompassing an object. Apart from CNNs, Transformers~\cite{wolf2020transformers} are also now being used in 2D object detection and it has shown promising outcomes. Carion et al.~\cite{carion2020end} formulates the 2D object detection problem by a direct set prediction approach. Similar to CornerNet, this also does not require anchor boxes. First, a CNN architecture generates 2D feature matrix that represents an input image. After flattening, this feature matrix is augmented with positional encoding. This is passed through a encoder-decoder transformer framework that proffers a set of output embeddings corresponding to object queries. A shared feed-forward network predicts the class-label and object bounding-box, or determines if the box represents no object inside it. The whole architecture is trained end-to-end with bipartite matching loss. 

\subsubsection{Region Proposal-based Two-Stage Architectures for 2D Object Detection}

In region-proposal based architectures, object detection is done in two stages. The first stage is class-agnostic where multiple region-proposals are generated from an image, without any regard towards the class to which it can belong. Thereafter, as part of the second stage, a ConvNet is used to extract features from these proposals and thereby, classify them into object categories. 

The Region-based Convolutional Neural Networks (R-CNN), conceptualized by Girshick et al.~\cite{girshick2014rich}, was a major landmark in the history of CNN-based object detection. The method contains three stages. As part of first stage of the R-CNN, a large number of region proposal ($\sim$2000) are generated by selective search on the input image. These proposals are independent of the object class to which they belong. Next, in the second stage, features are extracted from each such region proposal through a CNN, pre-trained on the task of image classification. To ensure that the input region proposals are compatible with the square-shaped fixed input shape of CNN, they are warped to a compact bounding-box around. The CNN is fine-tuned on these region proposals for all object classes and one extra class indicating background class (i.e. region proposal contains none of the intended objects). Finally, as part of the third stage, there are separate binary (positive/negative) Support Vector Machines (SVM) trained independently for each object class. These SVMs take in the feature vector from the last layer of CNN for a region proposal, and scores the proposals. Through Non-Maximum Suppression (NMS), regions with IoU overlap above a certain threshold are discarded and remaining are preserved. Further, the CNN features are also used to train a regression network to rectify and tighten the precise location and size of the bounding-box. Although this was a seminal work in itself, one of its major disadvantages was its computation time-complexity. Performing selective search on each input image to generate $\sim$2000 region proposals and passing each of them through CNN to generate features is a time-consuming process. Further, the three stages are independent of each other and thus, do not share weights amongst them. This makes the overall object detection process slow. 

He et al.~\cite{he2015spatial} introduced Spatial Pyramid Pooling in CNN (SPPNet) to detect objects in images. In R-CNN, the last fully-connected layer of CNN necessitated the input image to be of fixed size. To do away with this, the authors added a spatial pyramid pooling layer after the final convolutional layer, such that the input to the fully-connected layer is of fixed dimension. The overall object detection process gets faster, because for every input image, the CNN needs to do a single pass. But, the training is again time-consuming and weights of convolutional layers prior to spatial pyramid pooling layer are not updated during fine-tuning the deep neural network architecture on the region proposals.

\begin{figure}[h!]
	\includegraphics[width=0.97\linewidth]{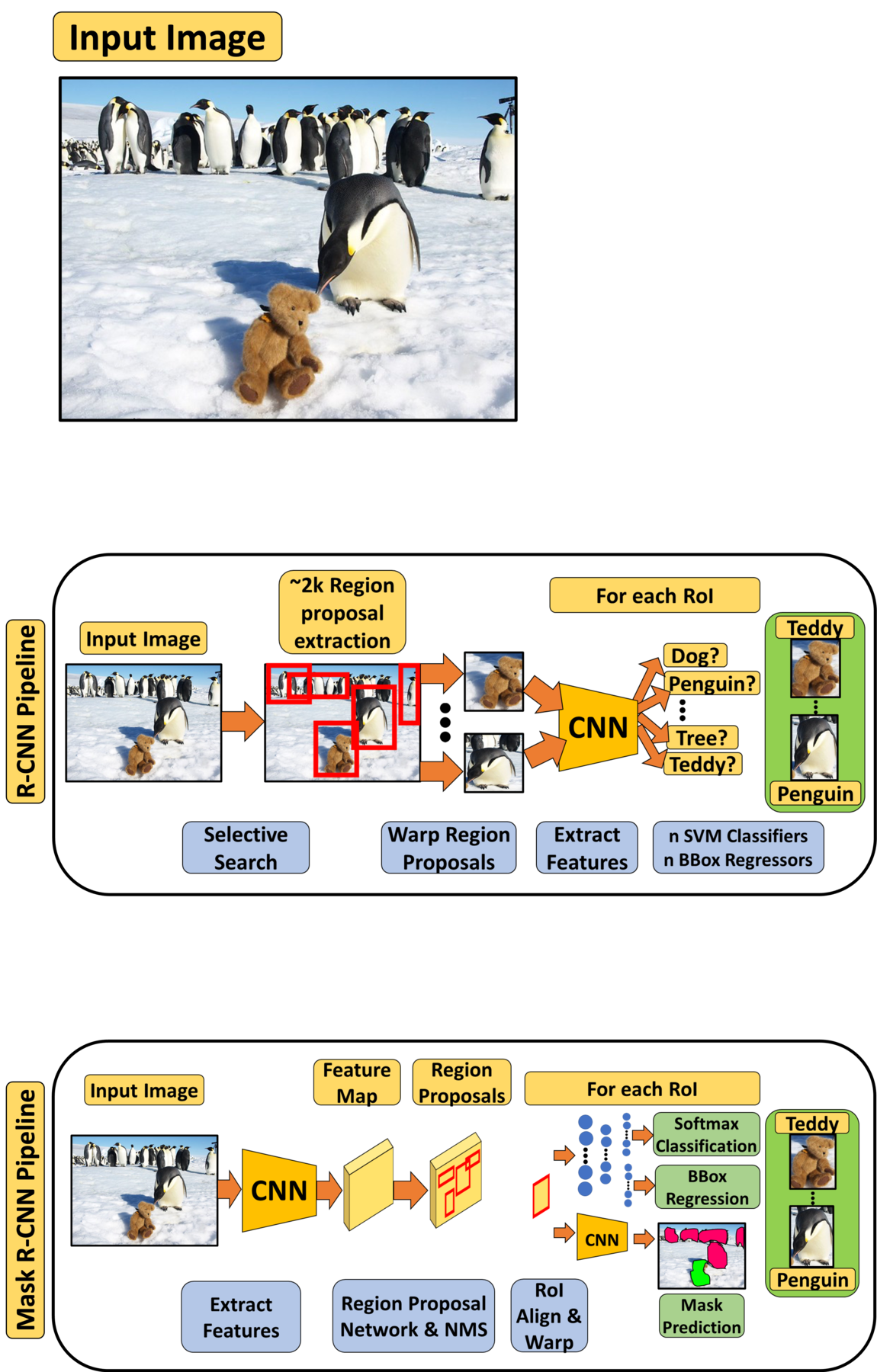}
	\caption{Comparison of R-CNN and Mask R-CNN pipeline (both two-stage architectures) for an image from COCO~\cite{lin2014microsoft}}
	\label{fig:rcnn_maskrcnn_pipeline}
\end{figure}

Girshick~\cite{girshick2015fast} later improved upon R-CNN and proposed Fast R-CNN. The idea is similar to R-CNN but here, instead of three separate stages, there is a unified process now with sharing of CNN weights. Unlike R-CNN, here the CNN does not take region proposals as input. Instead, the CNN accepts the raw RGB image as input. The CNN thereafter, through its multiple convolutional and max pooling layers, generates a convolutional feature map. Region proposals are generated from this convolutional feature map and a Region-of-Interest (RoI) Pooling generates a fixed-size feature vector for each such region proposal. A fully-connected layer accepts this feature vector and does two parallel predictions. The first is through a softmax layer corresponding to all object classes and one extra class indicating background. The second parallel prediction is to predict and refine the four corners of the bounding-box of region proposal. 

Another subsequent improvement by Ren et al.~\cite{ren2016faster} is through Faster R-CNN. Unlike R-CNN and Fast R-CNN, there is no separate selective search step here to generate object proposals. Instead, they use fully convolutional networks to learns generating region proposals. Keeping the proposals same, there is alteration between fine-tuning region proposal and object detection. Thereby, the weights of the CNN are shared between both these tasks and the process is even faster than earlier. 

Mask R-CNN was later proposed by He et al.~\cite{he2017mask}, as an extension on top of Faster R-CNN. We compare its framework with vanilla R-CNN in Figure~\ref{fig:rcnn_maskrcnn_pipeline}. As per the first stage of Mask R-CNN, a region proposal network generates region of interests in similar lines to Faster R-CNN. In the second stage, the bounding box and object category is predicted for each region of interest generated in the first stage. Along with this, Mask R-CNN has an additional parallel pipeline that generates object masks. These masks are similar to performing semantic segmentation~\cite{mukherjee2020two} on the whole image to generate pixel-wise classification. RoIPool serves extraction of feature map from each region of interest. But this may lead to quantization errors, which are handled by a RoIAlign layer (containing two convolutional layers) that aligns feature maps to the original input image. This in turn, generates pixel-level segmentation of each region of interest along with the original object detection pipeline of Faster R-CNN.

\section{Video Activity Recognition assisted by Object Detection}
\label{vidActbyObjDet}

In recent times, there is an abundance of object detection techniques, each outperforming its predecessors. As such, we have reached a stage where researchers are moving on to improve performance on even complex computer vision problems, viz. 3D object detection~\cite{misra2021end}, action detection and localization~\cite{jana2021unsupervised} tracking objects across videos, event recognition and scene understanding~\cite{jana2019multi}\cite{jana2019key}, etc. In this chapter, we focus on the task of event/activity recognition in videos, done with the assistance of frame-wise object detection, which enables inter-frame tracking of objects. This can be done in two ways broadly. The first way is to use per-frame object detection and thereafter, instance identification such as to track the same instances. Another way is to apply multi-frame fusion of information to identify same-instance tracking. This is particularly effective because every object shows minimal movement from one frame to the next. As such, there should be a significantly high Intersection-over-Union overlap of bounding boxes for the same object, in consecutive frames. 

Buri{\'c} et al.~\cite{buric2018object} employed object detection as a prerequisite for efficient activity recognition in sports videos. Specifically, they consider the team sports of handball and identify salient objects like players, ball, etc. Videos were recorded by GoPro camera in complex, challenging environments of sports hall/complex (for indoor) and playground (for outdoor) that exhibits varied artifacts like cluttering, shadows, low illumination and jitters. They compared three methods, viz. frame-wise CNN-based object detection like YOLOv2, Mask R-CNN and moving foreground-object detector like Mixture of Gaussians (MoG)~\cite{stauffer1999adaptive}. The principle behind MoG method is background subtraction, which is specifically useful in videos where the background is relatively static. By averaging the RGB values of all frames, we get an estimate of background. This background, when subtracted from each frame, proffers the foreground region in the frame. MoG is more robust towards non-static backgrounds, where history of each pixel is modeled through an approximation by K-Means clustering on the Expectation-Maximization algorithm. Results showed that YOLOv2 always predicted bounding-box around salient foreground objects, but had difficulty in distinguishing objects camouflaged with its immediate background. MoG was oftentimes erroneous in this task, because apart from real objects, it detected bounding-box around background patches due to illumination changes e.g. reflections, floodlights etc. Nevertheless, it did not face issues in camouflaging objects as it is driven by motion, not colors. Mask R-CNN was better in detecting smaller objects but similar to YOLOv2, it faced difficulties in detecting the ball (which was not an issue in MoG). However, none of the three methods were confused by shadows and did not end up detecting shadows as real objects. 

Ivasic-Kos et al.~\cite{ivasic2018building} suggests methodologies to track object instances across frames. This can be specifically helpful in video event detection too. They use Mask R-CNN to detect players in individual frames of handball videos. Thereafter, this detection and location information is combined with a measure of the players' activity by spatio-temporal interest points. This helps in tracking the same player throughout the entire video. 

Gleason et al.~\cite{gleason2019proposal} discussed action classification and spatio-temporal action detection in videos with the help of object detection. First, they perform frame-wise object detection by Mask R-CNN. These detection results are aggregated by hierarchical clustering approach whereby object detection bounding boxes are represented by a 3D coordinate comprised of frame number, x-coordinate and y-coordinate of the bounding-box center, and thereby clustered. Thereafter, by temporal jittering, dense action proposals are generated from the clusters proffered by hierarchical clustering. Finally, action classification was performed by training a Temporal Refinement I3D architecture that simultaneously classifies actions and refines the time-bounds for detected actions.

\section{Recent Datasets on 2D Object Detection}
\label{datasets}

Research in deep learning-based object detection flourished through the introduction of large-scale datasets containing fine-grained annotations of varied objects. In this regard, PASCAL VOC~\cite{everingham2015pascal} and MS COCO~\cite{lin2014microsoft} are two datasets that have been used extensively in various object detection research. PASCAL VOC contains $\sim$11,000 images spread over 20 categories which include person and  subcategories of animal (e.g., bird, horse), indoor objects (e.g., boat, train) and vehicles (e.g., sofa, tv). MS COCO is an even larger dataset with about 328,000 images spread over 91 object categories, which include the PASCAL VOC classes and also some rarely occurring categories such as toothbrush, toaster, wine glass, etc.  

More datasets are being introduced regularly by the research community, that are even larger in terms of scale and cater to detection of both  general objects and specialized objects such as guns, camouflaged objects, etc. We now present some of the datasets for generalized and specialized object detection in images, which have gained attention in recent times:

\vspace{0.3cm}

\noindent\textbf{Dataset for Object deTection in Aerial Images (DOTA)~\cite{ding2021object}.} This is a large-scale dataset intended for the task of detecting objects in aerial images. In its version 2.0, there are 11,268 images covering about 1.8 million object instances. Images are collected mainly from Google Earth, CycloMedia and other organizations that collect satellite and aerial images. Object categories range across 18 classes involving \textit{transportation} like large or small vehicles, ships helicopters, \textit{architectures} like bridge, harbors, \textit{ground patches} like tennis courts, swimming pool, helipad, etc.

\vspace{0.3cm}

\noindent\textbf{Objects in Context in Overhead Imagery (xView)~\cite{lam2018xview}.} This dataset is for object detection tasks in complex overhead scenes, in the form of images. The aim of such a dataset is to support models that can identify areas hit hard by catastrophes and disasters and thereby can speed-up the response and disaster management. Images are in the form of satellite images taken from  WorldView-3 satellites. In total, the images contain about 1 million objects spread over 60 object categories. Objects range from commonly spotted real-life instances like cars, buildings to land-use locations like vehicle lot, construction site, helipad and large objects like bus, trailer, tank car, etc. 

\vspace{0.3cm}

\noindent\textbf{Object365~\cite{shao2019objects365}.} Object365 is a dataset focused towards supporting detection of a diverse range of objects in the wild. This dataset contains around 30 million object instances spread over 2 million images. The number of object classes is 365, which can be grouped under 11 high-level categories. The high-level categories include \textit{human \& accessory} (e.g. person, watch, book), \textit{clothes} (e.g. slippers, mask, hat), \textit{kitchen items}  (e.g. plate, refrigerator, knife), \textit{animals}  (e.g. crab, yak, dolphin), \textit{food}  (e.g. carrot, avocado, coconut), \textit{electronics}  (e.g. printer, microphone, compact disk), etc.

\vspace{0.3cm}

\noindent\textbf{Open Images Dataset~\cite{kuznetsova2020open}.} The dataset comprises almost 16 million object instances on 1.9 million images. The objects can be grouped into 600 object classes. The objects cover a varied range of real-world entities. Objects include course-grained categories (e.g. animals), fine-grained categories (e.g. shellfish), scenes, events (e.g. birthday, wedding), materials, etc.

\vspace{0.3cm}

\noindent\textbf{Camouflaged Object Detection (COD10k)~\cite{fan2020camouflaged}.} This dataset is composed of images collected from photography websites (e.g. Flickr), where the intended objects to be detected are obscured organisms within its surroundings. There are around 10,000 images, with annotated object bounding-boxes. The objects can be categorized into 78 classes (out of which 69 classes are of camouflaged objects, 9 classes are of non-camouflaged objects), which can be again grouped into 10 super-classes (out of which 5 super-classes are of camouflaged objects, rest are of non-camouflaged objects). Object examples include dog, fish, mantis, toad, deer, cicada, etc., which are grouped into super-classes like aquatic animals, flying animals, terrestrial animals, amphibians and others.

\vspace{0.3cm}

\noindent\textbf{Large Vocabulary Instance Segmentation (LVIS)~\cite{gupta2019lvis}.} The dataset contains approximately 2 million object instance annotations across 0.16 million images. The images were mostly obtained from the COCO dataset~\cite{lin2014microsoft}. Objects are categorized into 1000 categories of regular everyday objects e.g. bed, tea-cup, peanut, donuts, umbrella, shoulder bag.

\vspace{0.3cm}

Further, we elaborate on some recent datasets for object detection in video clips:

\vspace{0.3cm}

\noindent\textbf{YouTube-BoundingBoxes (YT-BB)~\cite{real2017youtube}.} Unlike the aforementioned datasets, this dataset focuses on object detection in videos. It comprises about 5.6 million object instances (annotated in the form of bounding box) across 380,000 trimmed videos of approximately 15-20 seconds duration, which specifically features objects appearing in natural scenes. Videos were sourced from YouTube. Video objects are categorized into 23 object classes, encompassing, person, animals (e.g. dog, elephant) or other common objects (e.g. aeroplane, crowd).  

\vspace{0.3cm}

\noindent\textbf{Temporal Hands Guns and Phones Dataset (THGP)~\cite{duran2021tyolov5}.} This is also a dataset catering to object detection in videos. In the training partition, there are 50 videos, each of 100 frames. The test partition, on the other hand, contains 48 videos of 20 frames each. The ongoing events in the videos are specific to policing and crimes e.g. armed robberies in stores, shooting drills, making calls. As such, the objects like hands, guns, phones are annotated in the video frames. 

\vspace{0.3cm}

\noindent\textbf{Objects Around Krishna
	(OAK)~\cite{wang2021wanderlust}.} This dataset comprises of 80 video clips spanning about 17 hours. Bounding-box annotations are provided for detecting objects continually, in the video frames. Videos are egocentric in nature and were captured by graduate students over a long period of time by a camera called KrishnaCam. Since the videos represent regular outdoor scenes, the objects are effectively natural everyday objects, which are categorized into 105 object classes. 

\begin{table*}[!h]
	\centering
	\caption{Performance (mAP) of state-of-the-art 2D object detection methods in images, on some of the widely used benchmark datasets}
	\resizebox{0.98\textwidth}{!}{%
		\begin{tabular}{c|c||c|c}
			\hline
			\textbf{Method} & \textbf{CNN Backbone} &
			\textbf{PASCAL VOC~\cite{everingham2015pascal}} & 
			\textbf{MS COCO~\cite{lin2014microsoft}}%
			\TstrutNorm\BstrutNorm\\ \hline
			\textbf{\textit{Monolithic F/w}} &   &   &  \TstrutNorm\\
			SSD513~\cite{liu2016ssd} & ResNet-101 & 76.8 &  31.2\\
			DSSD513~\cite{fu2017dssd} & ResNet-101 & 81.5 &  33.2\\
			YOLO~\cite{redmon2016you} &  $\sim$GoogLeNet & 66.4  &  -\\
			YOLOv2~\cite{redmon2017yolo9000} & DarkNet-19 & 78.6 &  21.6\\
			YOLOv3~\cite{farhadi2018yolov3} & DarkNet-53 & - &  33.0\\
			YOLOv4~\cite{bochkovskiy2020yolov4} & CSPDarkNet-53 & - &  43.5\\
			CornerNet511~\cite{law2018cornernet} & Hourglass-104 & - &  42.1\\
			RetinaNet~\cite{lin2017focal} & ResNet-101 & - &  39.1\\
			EfficientDet-D7~\cite{tan2020efficientdet} & BiFPN & - &  52.2\\
			DETR-DC5~\cite{carion2020end} & ResNet-101 & - &  44.9\\
			Swin Transformer~\cite{liu2021swin} & HTC++ & - &  58.7\\\hline
			\textbf{\textit{Two-Stage F/w}} &   &   &  \TstrutNorm\\
			SPP-Net~\cite{he2015spatial} & ZF-5 & 60.9  &  -\\
			R-CNN~\cite{girshick2014rich} & AlexNet & 58.5  &  -\\
			Fast R-CNN~\cite{girshick2015fast} & VGG-16 & 70.0  & 19.7 \\
			Faster R-CNN~\cite{ren2016faster} & VGG-16 & 73.2  & 21.9 \\
			FPN~\cite{lin2017feature} & ResNet-101 & - & 36.2 \\
			Mask R-CNN~\cite{he2017mask} & ResNeXt-101 & -  & 39.8\\
			Cascade R-CNN~\cite{cai2019cascade} & ResNeXt-101 & -  & 45.8\\
			DetectoRS~\cite{qiao2021detectors} & ResNeXt-101 & -  & 55.7\BstrutNorm\\
			\hline
	\end{tabular}}
	
	\label{tab_sota}
\end{table*}

\section{Performance Comparison of Object Detection Techniques}
\label{perfComparison}

In Table~\ref{tab_sota}, we compare the performance of some of the state-of-the-art object detection techniques on the PASCAL VOC~\cite{everingham2015pascal} and MS COCO~\cite{lin2014microsoft} datasets. As evident from Table~\ref{tab_sota}, PASCAL VOC was mostly used for experimentation in the earlier research papers. Its predominance was later replaced by MS COCO dataset, which contains more object annotations and is a superset of all the PASCAL VOC object categories. For PASCAL VOC, the metric used for comparison is  interpolated average precision. The threshold of Intersection-over-Union (IoU) for predicted and ground-truth bounding boxes is 0.5. In case of MS COCO, the metric used for comparison is mean Average Precision (mAP) performance, which is computed by averaging performance over ten values of IoUs in the range [0.5, 0.95] at intervals of 0.05. 

\section{Conclusion and Way Forward}
\label{epilogue}
In this chapter, we elaborate upon some of the recent state-of-the-art object detection techniques in use. Monolithic architecture exhibit a single unified ConvNet and/or Transformer pipeline for bounding-box regression and object classification. Two-stage architectures have a separate region-proposal step whose extracted features are used to predict object category and bounding-box coordinates. We compare and contrast some of the significant monolithic and two-stage architecture based object detection techniques. We also discuss the usefulness of using frame-wise object detection in videos for the task of activity recognition and video event classification. Finally, the chapter details some of the recent large-scale datasets for 2D object detection in images and videos, that can be used to train deep neural network based object detection architectures. 

In future, trends of object detection would most likely include more few-shot or zero-shot learning approaches to do away with the need for large-scale annotated datasets, catering to both commonly- and rarely-occurring object categories. Researchers are also moving towards more explainable object detection approaches to avoid inadvertent false negatives in critical emergency deployments like 24$\times$7 surveillance, rescue operation etc. For applications in autonomous vehicles and indoor scene understanding, 3D object detection in point clouds is another emerging research area, that is used to recognize and localize objects in 3D environments.

%\nocite{*} % add all entries from sample.bib

\bibliography{ref.bib}

\end{document}